\documentclass[11pt]{article}

\usepackage[]{acl}

\usepackage{times}
\usepackage{latexsym}

\usepackage[T1]{fontenc}

\usepackage[utf8]{inputenc}

\usepackage{microtype}

\usepackage{inconsolata}

\usepackage{graphicx}

\usepackage[ruled,vlined]{algorithm2e}
\usepackage{float}                        
\usepackage{amsmath}
\usepackage{amsthm}
\newtheorem{definition}{Definition}
\usepackage{listings}
\usepackage{makecell}
\usepackage[most]{tcolorbox}
\usepackage{enumitem}
\usepackage[normalem]{ulem}
\usepackage{xcolor}   
\usepackage{pifont}   
\newcommand{\meanstd}[2]{#1 {\scriptsize\normalfont$\pm$ #2}}
\usepackage{float} 
\usepackage{multirow}

\title{Thinking Before Constraining: A Unified Decoding Framework for Large Language Models}

\author{
\textbf{Ngoc Trinh Hung Nguyen\textsuperscript{1,2}}, 
\textbf{Alonso Silva\textsuperscript{2}}, 
\textbf{Laith Zumot\textsuperscript{3}} 
\\
\textbf{Liubov Tupikina\textsuperscript{2}}, 
\textbf{Armen Aghasaryan\textsuperscript{2}}, 
\textbf{Mehwish Alam\textsuperscript{1}}
\\
\\
\textsuperscript{1} Télécom Paris, Institut Polytechnique de Paris, France, 
\textsuperscript{2} Nokia Bell Labs, 
\textsuperscript{3} Nokia; Independent Researcher
\\
\\
\small{
\textbf{Correspondence:} \href{mailto:ngoc.nguyen@ip-paris.fr}{ngoc.nguyen@ip-paris.fr}
}
}

\begin{document}
\maketitle
\begin{abstract}
Natural generation allows Large Language Models (LLMs) to produce free-form responses with rich reasoning, yet the lack of structure makes outputs difficult to verify. Conversely, constrained decoding ensures standardized formats but can inadvertently restrict reasoning capabilities by imposing constraints too early in the generation process. We propose a hybrid approach, namely \emph{In-Writing}, that combines free-form reasoning and structured generation in a single call. The model first performs unconstrained reasoning and only applies structured decoding after a trigger token is generated, explicitly decoupling reasoning from formatting. We establish that our trigger-token strategies are able to virtually eradicate \textit{premature triggering}, a failure mode in which constrained decoding interrupts ongoing reasoning. Evaluations across diverse datasets covering classification and reasoning tasks demonstrate that our approach outperforms the state-of-the-art by achieving accuracy gains of up to 27\% over natural generation. Our code are available at:
\url{https://github.com/Nokia-Bell-Labs/InWriting}. \end{abstract}

\section{Introduction}

\begin{figure}[h!]

\begin{tcolorbox}[
  colback=gray!10,
  colframe=black,
  boxrule=0.5pt,
  arc=2mm,
]
Question: Alice, Bob, Claire, Dave, and Eve are dancers at a square dance. At the start of a song, they each have a partner: Alice is dancing with Helga, Bob is dancing with Ophelia, Claire is dancing with Sam, Dave is dancing with Melissa, and Eve is dancing with Lola. Throughout the song, the dancers often trade partners. First, Eve and Claire switch partners. Then, Bob and Eve switch partners. Then, Dave and Bob switch partners. Then, Alice and Bob switch partners. Finally, Alice and Claire switch partners. At the end of the dance, Bob is dancing with:

A) Helga. B) Ophelia. C) Sam. D) Melissa. E) Lola.

Answer: A
\end{tcolorbox}

\begin{tcolorbox}[
  colback=red!5,
  colframe=black,
  boxrule=0.5pt,
  arc=2mm,
]

\small
\textbf{Natural Generation: }

Let's break down the sequence of partner switches step by step:\\
1. Eve and Claire switch partners: Eve is now dancing with Sam, and Claire is dancing with Lola.\\
2. Bob and Eve switch partners: Bob is now dancing with Lola, and Eve is dancing with Melissa.\\
3. Dave and Bob switch partners: Dave is now dancing with Lola, and Bob is dancing with Melissa.\\
4. Alice and Bob switch partners: Alice is now dancing with Melissa, and Bob is dancing with Helga.\\
5. Alice and Claire switch partners: Alice is now dancing with Lola, and Claire is dancing with Helga.\\
At the end of the dance, Bob is dancing with... Helga!

\hfill {\Large \textcolor{red}{\ding{55}}}

\end{tcolorbox}

\begin{tcolorbox}[
  colback=green!8,
  colframe=black,
  boxrule=0.5pt,
  arc=2mm,
]
\small
\textbf{In-Writing:}

Let’s break down the sequence of partner switches step by step:\\
1. Eve and Claire switch partners: Eve is now dancing with Sam, and Claire is dancing with Lola.\\
2. Bob and Eve switch partners: Bob is now dancing with Lola, and Eve is dancing with Melissa.\\
3. Dave and Bob switch partners: Dave is now dancing with Lola, and Bob is dancing with Melissa.\\
4. Alice and Bob switch partners: Alice is now dancing with Melissa, and Bob is dancing with Helga.\\
5. Alice and Claire switch partners: Alice is now dancing with Lola, and Claire is dancing with Helga.\\
At the end of the dance, Bob is dancing with... Helga! \\
\{  ``final\_answer'': ``A'' \}

\hfill {\Large \textcolor{green}{\ding{51}}}

\end{tcolorbox}

\caption{Comparison between Natural Generation (NL) and the proposed In-Writing method on Llama3-8B-it. Both produce identical reasoning; In-Writing additionally generates an easily parsable final answer with syntactic corrections (e.g., mapping ``Helga'' to choice \textbf{A}.).
}

\label{fig:natural_vs_structured}
\end{figure}

Large Language Models (LLMs) have demonstrated remarkable capabilities across a wide range of applications, including text completion, summarization, question answering, code generation, web navigation, data extraction, and tool use \citep{zhao2023survey, minaee2024large}. Trained primarily on large natural language corpora, LLMs typically generate fluent and flexible text at inference time. As a consequence, they do not inherently guarantee adherence to predefined output structures. Although advanced LLMs often produce syntactically well-formed outputs, this behavior is not assured \citep{koo2024automata}. The absence of strict structural guarantees can limit the applicability of LLMs in tasks such as schema-based information extraction, structured question answering, and many industrial use cases \citep{liu2024we}.

To address structured output requirements, grammar-constrained decoding methods \citep{willard2023efficient,guidance2025,dong2024xgrammar} restrict token generation by masking invalid continuations, ensuring syntactic correctness. However, these approaches may reduce expressiveness and fluency, harming generalization \citep{tam-etal-2024-speak, banerjee2025crane, lee2026format}.

A common alternative, \textbf{NL-to-Format}, employs a two-stage pipeline where a model first generates Natural Language (NL) and a second model converts it into the target format \citep{tam-etal-2024-speak, lee2026format}. This approach improves flexibility but increases computational cost and redundancy, without guaranteeing adherence to predefined output structures \citep{lundberg2023art, lee2026format}.

Another approach, \textbf{CRANE}, alternates between free-form and constrained generation using delimiter-based (\texttt{<< >>}) switching, where reasoning is enclosed within delimiters \citep{banerjee2025crane}. While improving parsability, CRANE relies on complex prompting and is typically limited to single-field formats, and reintroduces concerns about reasoning under constrained decoding.



To bridge this gap, we introduce \textbf{In-Writing}, a framework that unifies natural and structured generation by allowing reasoning to proceed in free form until specific trigger tokens (e.g., \texttt{<eos>} or \texttt{\{}) activate structured decoding. Unlike prior methods that tightly couple reasoning with constrained regions (e.g., CRANE) or rely on rigid pipelines (e.g., NL-to-Format), our approach decouples reasoning from formatting. We further study trigger-token strategies to mitigate \textit{premature triggering}, a failure mode in which constrained decoding interrupts ongoing reasoning.

We evaluate our framework across diverse downstream tasks and model families, including Qwen, Llama, Gemma, DeepSeek and SmolLM models (1.5B–14B parameters). Our evaluation spans classification and reasoning benchmarks, using accuracy, parsability, and token efficiency as metrics to demonstrate effectiveness.

Our main contributions are: (1) a framework that combines natural and structured generation with minimal token overhead; (2) evidence that constrained decoding serves as a more effective parser and corrector than using separate models to extract final answers (see Figure~\ref{fig:natural_vs_structured}); (3) an investigation of trigger-token strategies to mitigate the premature triggering problem; and (4) an empirical demonstration that forcing models to reason entirely within a constrained grammar space is less effective than our framework.

This paper is structured as follows: Section~\ref{sec:preliminaries} introduces preliminary work on Finite Automata. Section~\ref{sec:related_work} reviews related work while Sections~\ref{sec:in_writing_method}, \ref{sec:experiments}, and \ref{sec:results_discussion} present our method and experimental results respectively, demonstrating its effectiveness and robustness.

\section{Preliminaries}\label{sec:preliminaries}

We first introduce the formal definition of a finite automaton, also known as a Finite-State Machine (FSM) \citep{sipser1996introduction}, which is one of the central components in constrained decoding \citep{willard2023efficient, koo2024automata}, as they can represent a target regular expression (or regex)\footnote{We use the mathematical definition of regular expression ~\citep[Def.~1.52, p.~64]{sipser1996introduction}.} and guide the model to select only tokens that are consistent with that regex at each decoding step.



\begin{definition}[Finite Automaton]\label{def:finite_automaton}
A finite automaton, or finite-state machine is a 5-tuple $(Q, \Sigma, \delta, q_0, F)$, where
$Q$ is a finite set of states, $\Sigma$ a finite alphabet,
$\delta: Q \times \Sigma \to Q$ the transition function,
$q_0 \in Q$ the start state, and $F \subseteq Q$ the set of accept states.
\end{definition}

\section{Related Work}\label{sec:related_work}


Token generation in language models follows an autoregressive process, where each token is sampled from the conditional distribution given previous tokens \citep{bengio2003neural}. However, standard sampling is stochastic and does not enforce structural constraints, limiting its applicability in settings requiring strict output formats \citep{scholak-etal-2021-picard,geng2023grammar,tam-etal-2024-speak}. Tool calling enables LLMs to invoke external tools and APIs \citep{schick2023toolformer, yao2022react}, and has also been used for structured output generation (e.g., JSON) \citep{instructor2023}. However, since tool calls are generated autoregressively, strict structural correctness is not guaranteed.


Hard constrained decoding is yet another technique which enforces structured generation by applying decoder logit masking during sampling \citep{zhang2019neural, deutsch2019general}, as shown in Algorithm~\ref{alg:masked_generation}. Recent work formulates regex- or grammar-guided generation as a finite-state machine (FSM) (Definition~\ref{def:finite_automaton}), enabling controllable initialization and termination \citep{willard2023efficient, koo2024automata}. However, hard constrained decoding has been shown to degrade LLM performance on certain tasks \citep{tam-etal-2024-speak, banerjee2025crane, lee2026format}.

\begin{algorithm} 
\caption{LM Token Sampling with Masking} 
\label{alg:masked_generation}

\SetAlgoLined
\LinesNumbered
\SetKwFunction{FMain}{sample\_tokens\_masked}
\FMain{$rx, p, L$} \tcp*{$rx$: regex or grammar, $p$: prompt, $L$: max new tokens}

$s \gets (p)$ \tcp*{Generated token sequence} 
\For{$i \gets 1$ \KwTo $L$}{ 
    $\alpha \gets LM(s)$\; 
    Construct mask $m(s, rx)$\; 
    $\tilde{\alpha} \gets m \odot \alpha$\; 
    Sample $\tilde{t}$ from $\tilde{\alpha}$ \;
    \If{$\tilde{t} = \text{EOS}$}{ break\; } 
    $s \gets \text{append}(s, \tilde{t})$\; 
} 
\Return $s$\; \end{algorithm}

A few recent works have introduced hybrid approaches such as  \textbf{NL-to-Format} and \textbf{CRANE}. \textbf{NL-to-Format}  uses a two-stage pipeline in which a model first generates an answer in natural language and a second model converts it into the target format \citep{tam-etal-2024-speak, lee2026format}. \textbf{CRANE} interleaves free-form and grammar-constrained generation via delimiter-based (\texttt{<< >>}) switching \citep{banerjee2025crane}. However, requiring constrained fields to remain consistent across invocations can hinder multi-field or complex extraction with heterogeneous constraints. Moreover, the final response may not be reliably extractable, as there is no guarantee it will be generated within \texttt{<< >>} beyond prompting. This also reintroduces concerns regarding reasoning under constrained decoding.

Our work addresses the gap between reasoning expressiveness (limitation of hard-constrained decoding and CRANE, which restrict free-form reasoning), format guarantees (limitation of natural generation and NL-to-Format, which lack strict structural enforcement), and computational overhead, by proposing a unified hybrid constrained decoding framework that preserves reasoning flexibility while ensuring strict format compliance with minimal overhead.

\section{In-Writing Method}\label{sec:in_writing_method}





In this section, we introduce \textbf{In-Writing}, a decoding framework that separates reasoning from structural formatting constraints. We first formalize our method in Section~\ref{sec:prob_formula}, followed by its operational decoding algorithm in Section~\ref{sec:algo}.

\subsection{Decoupled Probabilistic Formulation}\label{sec:prob_formula}

Given a question $q$, Chain-of-Thought (CoT) generation models the answer $A$ as being produced via latent reasoning traces $R$ \citep{wei2022chain} which is formulated as:
\begin{equation}
P(A \mid q) = \sum_{R} P(A \mid R, q) P(R \mid q).
\label{eq:CoT_decoding}
\end{equation}

When a formatting constraint $F$ is imposed via logit masking, conventional constrained decoding conditions both reasoning and answer generation on $F$ as follows:
\begin{equation}
P(A \mid q, F) = \sum_{R} P(A \mid R, q, F) P(R \mid q, F).
\label{eq:hard_constrained_full_formula}
\end{equation}

Under hard-constrained decoding, invalid reasoning paths are removed from the search space, as formalized in the following equation:
\begin{equation}
P(R \mid q, F) =
\frac{1}{Z} P(R \mid q)\,\mathbf{1}_{R \in F},
\label{eq:hard_constrained}
\end{equation}
where $\mathbf{1}_{R \in F}$ is an indicator function enforcing compliance with $F$ (1 if $R \in F$, 0 otherwise), and $Z$ is a normalization. Consequently, logically valid reasoning traces may be discarded due to minor structural violations.

To address this limitation, In-Writing delays formatting until reasoning is complete, effectively decoupling reasoning from formatting constraints and making it independent of the formatting condition, as formalized below: 
\begin{equation}
P(R \mid q, F) = P(R \mid q).
\label{eq:InWriting_constrained}
\end{equation}

This independence enables reformulating Equation~\ref{eq:hard_constrained_full_formula} as follows:
\begin{equation}
P(A \mid q, F) = \sum_{R} P(A \mid R, q, F) P(R \mid q).
\label{eq:InWriting_constrained_full_formula}
\end{equation}

Equation~\ref{eq:InWriting_constrained_full_formula} shows that In-Writing retains both the reasoning expressiveness of CoT decoding (Equation~\ref{eq:CoT_decoding}) and the formatting guarantees of hard-constrained decoding (Equation~\ref{eq:hard_constrained_full_formula}).


\subsection{State-Based Decoding Algorithm}\label{sec:algo}

In-Writing is instantiated in Algorithm~\ref{alg:Text2Schema_generation}, where LLMs first generate an unconstrained reasoning trace (state~$-1$; lines~1--6) until a trigger token is emitted, then switch to structured generation (state~$0$; lines~8--16). In this way, constrained decoding is applied only after reasoning is completed, effectively acting as a parser that converts the final reasoning trace into a structured output. An illustration is given in Section~\ref{sec:illustration_in_writing}.

\begin{algorithm} 
\caption{LM Token Sampling with In-Writing Method} 
\label{alg:Text2Schema_generation}
\SetAlgoLined
\LinesNumbered
\SetKwFunction{FMain}{sample\_InWriting}
\FMain{$rx, triggers, p, L$} \tcp*{$rx$: regex or grammar, $triggers$: trigger tokens, $p$: prompt, $L$: max new tokens}

$s \gets (p)$ \tcp*{Generated token sequence} 

\For{$i \gets 1$ \KwTo $L$}{ 
    $\alpha \gets LM(s)$\;
    Sample $t$ from $\alpha$\; 
    \If{$t \notin \text{triggers}$}{
        $s \gets \text{append}(s, t)$\; 
    }
    \Else{
        \While{rx is not completed}{
            Construct mask $m(s, rx)$\;
            $\tilde{\alpha} \gets m \odot \alpha$\;
            Sample $\tilde{t}$ from $\tilde{\alpha}$\;
            $s \gets \text{append}(s, \tilde{t})$\;
        }
        \Return $s$\;
    }
} 
\Return $s$\; 
\end{algorithm}

This approach offers four key advantages: \textbf{(1) Unified pipeline:} it decouples reasoning and formatting within a single call, eliminating external verification. \textbf{(2) Guaranteed syntactic validity:} regex- or grammar-based constraints enforce schema-compliant outputs during decoding. \textbf{(3) Robustness:} outputs are reliably mapped to structured formats without complex prompting or implementation. \textbf{(4) Minimal overhead:} the formatting step introduces negligible latency.

We term this approach \textit{In-Writing}, drawing an analogy to diffusion-based \textit{inpainting} \citep{lugmayr2022repaint}, where generation is restricted to masked regions. Similarly, \textit{In-Writing} separates reasoning from formatting by allowing free-form generation while enforcing strict syntactic constraints only on target output slots.


\section{Experimental Setup}\label{sec:experiments}

This section describes the experimental setup used to evaluate our approach. All experiments were conducted on NVIDIA A40 GPU.

\subsection{Datasets}

Following the evaluation settings from \citep{tam-etal-2024-speak, banerjee2025crane}, we evaluate our hybrid approach on a diverse set of reasoning and classification benchmarks spanning numerical, symbolic, and textual outputs, using the same preprocessing and data splits as prior works.

\subsubsection{Reasoning Tasks}

We use the following datasets for the reasoning tasks.

\noindent \textbf{GSM8K} \citep{cobbe2021training} contains grade-school math problems which require multi-step reasoning.

\noindent \textbf{GSM-Symbolic} \citep{mirzadeh2025gsm} is a symbolic variant of GSM8K replacing numbers with variables to test compositional and algebraic generalization.

\noindent \textbf{Last Letter Concatenation} \citep{wei2022chain} is a symbolic task where the model concatenates the last letters of words.

\noindent \textbf{Shuffled Objects} \citep{ghazal2013bigbench} is a BigBench task requiring prediction of object arrangements after sequential shuffling operations.

\subsubsection{Classification Tasks}
We use the following datasets for evaluating over the classification tasks.

\noindent \textbf{DDXPlus} \citep{fansi2022ddxplus} is a 49-class medical diagnosis task \citep{wu2024streambench}.

\noindent \textbf{MultiFin} \citep{jorgensen2023multifin} is a 5-class financial text classification task.

\noindent \textbf{Sports Understanding} \citep{ghazal2013bigbench} is a BigBench task classifying sports-related sentences as plausible or implausible.

\noindent \textbf{NI - Task 280} \citep{mishra2022cross} is a multiple-choice stereotype classification task based on a given paragraph, which is highly sensitive to the format of the prompt \citep{sclar2023quantifying}.

\subsection{Evaluation Metrics}

We evaluate along two dimensions: correctness and parse success. Accordingly, we report \textbf{accuracy} and \textbf{parse rate} as the primary metrics.

\paragraph{Accuracy Metrics.}
For the Natural Language (NL) baseline, answers are extracted from free-form outputs using predefined prefixes (e.g., ``answer is:'') specified in the prompt. 

For NL-to-Format, a larger LLM extracts the final answer from the first-stage NL output. For Constrained Decoding and In-Writing, answers are parsed directly from the structured output. All extracted answers are evaluated against the ground-truth answer using exact string match.

\paragraph{Parse Metrics.}
For the NL baseline, parsability is defined by the presence of the expected answer prefix (e.g., ``answer is:''), followed by a syntactically valid extracted output.

For NL-to-Format, constrained decoding, and In-Writing, parsability is defined by whether the generated output is syntactically valid.

\paragraph{In-Writing Robustness Evaluation.}
In-Writing separates reasoning and formatting via trigger tokens, making it sensitive to trigger selection. We evaluate robustness across different trigger token sets. We denote \textbf{In-Writing-Base} as using \texttt{trigger\_token\_ids} = \{\texttt{<eos>}, \texttt{\{}\}, and \textbf{In-Writing*} as using \texttt{<eos>} as the sole trigger token. This setup enables analysis of \textit{premature triggering}, where early constraint activation truncates reasoning, and whether using a single \texttt{<eos>} trigger mitigates this issue.


Since both In-Writing and NL-to-Format extend the NL baseline, we additionally perform an \textbf{overlap analysis} to compare agreements and discrepancies across methods in Section~\ref{subsubsec:overlap_nl_format_inwriting}.

\subsection{Models and Prompts}
We evaluate 18 open-source models released between 2024 and 2026 from five families: Qwen, Llama, Gemma, DeepSeek, and SmolLM. Model sizes range from 1.5B to 32B parameters, all run locally using the \textit{Transformers} library.

\paragraph{NL-to-Format Comparison}
For direct comparability with \citet{tam-etal-2024-speak}, we adopt their prompt templates and baseline models: LLaMA3-8B-Instruct \citep{dubey2024llama} and Gemma2-9B-Instruct \citep{team2024gemma}. We extend the evaluation with newer models, including Qwen3 (1.7B, 4B, 8B) \citep{yang2025qwen3}, Qwen3.5 (2B, 4B, 9B) \citep{qwen3.5}, and SmolLM3-3B \citep{bakouch2025smollm3}.

\paragraph{CRANE Comparison}
Following \citet{banerjee2025crane}, we use their prompt templates with a small additional instruction (see Appendix~\ref{sec:gsm-symbolic}.), across open-source models: Qwen2.5 (1.5B, Coder-7B, Math-7B, Coder-14B) \citep{qwen2025qwen25technicalreport}, LLaMA3.1-8B-Instruct \citep{dubey2024llama}, and DeepSeek-R1 Distill variants (7B/14B Qwen, 8B LLaMA) \citep{guo2025deepseek}.

We retain the original benchmark templates to evaluate model performance under fixed-format prompts. The NL-to-Format benchmark requires natural-language prefixes (e.g., ``answer is:''), whereas CRANE enforces delimiter-based outputs (e.g., \texttt{<< >>}). These templates place our approach at a disadvantage by providing no guidance on the In-Writing output format. This setup tests whether models can map internal reasoning to required output formats without additional prompt engineering.

The prompt template is provided in Appendix~\ref{sec:prompt-template}.

\subsection{Output Schema}

Although regex or grammar constraints can be specified independently, we use Pydantic\footnote{\url{https://docs.pydantic.dev/latest/}} to automatically define and customize JSON schemas. For vanilla constrained decoding, outputs are strictly formatted as JSON with two keys: \texttt{think\_step\_by\_step} for reasoning and \texttt{final\_answer} for the answer, whereas In-Writing first generates free-form reasoning before emitting the final JSON, from which the answer is extracted via the \texttt{final\_answer} field. While our approach is not limited to JSON, we focus on JSON schemas due to their simplicity and ease of conversion to formats such as XML or YAML.

\subsection{Experimental Implementation Details}

\paragraph{NL-to-Format Comparison.}
We report Baseline (NL) results by extracting the natural-language answer from LLMs using the \textbf{``answer is:''} prefix, as specified in the prompt template.

For NL-to-Format results, the complete first-stage output (i.e., Baseline (NL)) is provided as input to a larger second-stage parser model. For reproducibility, we use the open-source Qwen3-32B instead of \textit{claude-3-haiku-20240307} used in prior work~\citep{tam-etal-2024-speak}.

Vanilla constrained decoding (\textbf{Constrained}) and In-Writing are implemented using Litelines~\citep{litelines}, an open-source library built on top of Outlines~\citep{willard2023efficient}. Litelines extends Outlines with \texttt{allow\_preamble} and \texttt{trigger\_token\_ids}, enabling the model to reason freely outside the schema until a trigger token is generated.

\paragraph{CRANE Comparison.} We evaluate only the In-Writing, as Baseline, Constrained, and Chain-of-Thought results are already reported by \citet{banerjee2025crane}.

\section{Results and Discussion}\label{sec:results_discussion}

In this section, we evaluate In-Writing against state-of-the-art (SOTA) methods. We compare In-Writing with two approaches: (1) a two-stage natural language-to-format (NL-to-Format) pipeline (Section~\ref{sec:NL-to-Format_comparison}) and (2) a hybrid constrained decoding approach (CRANE) (Section~\ref{sec:CRANE_comparison}).

\subsection{NL-to-Format Comparison: Structured Generation as an Effective Parser}\label{sec:NL-to-Format_comparison}

\subsubsection{Overall Assessment}

We first evaluate In-Writing-Base and In-Writing* on seven datasets (GSM8K, Last Letter Concatenation, Shuffled Objects, DDXPlus, MultiFin, Sports, and Task 280) using the same prompts, models (LLaMA3-8B-Instruct and Gemma2-9B-Instruct), and zero-shot setting as prior work \citep{tam-etal-2024-speak}; results are shown in Tables~\ref{tab:reasoning_base_lmsf_results} and~\ref{tab:classification_results}.

\begin{table}
  \centering
  \small
  \begin{tabular}{llll}
    \hline
    \textbf{Task} & \textbf{Model} & \textbf{Method} & \textbf{Acc (\%)} \\
    \hline

    \multirow{8}{*}{GSM8K}
    & \multirow{4}{*}{LLaMA}
      & Baseline (NL)       & \meanstd{66.2}{9.5} \\
    &  & Constrained     & \textit{\meanstd{48.9}{6.7}} \\
    &  & NL-to-Format    & \textit{\meanstd{74.7}{0.6}} \\
    &  & In-Writing*     & \textbf{\meanstd{77.9}{1.3}} \\
    \cline{2-4}

    & \multirow{4}{*}{Gemma}
      & Baseline (NL)        & \meanstd{82.7}{7.2} \\
    &  & Constrained     & \textit{\meanstd{84.2}{3.7}} \\
    &  & NL-to-Format    & \textit{\meanstd{86.5}{0.6}} \\
    &  & In-Writing*     & \textbf{\meanstd{86.9}{0.4}} \\
    \hline

    \multirow{8}{*}{Last Letter}
    & \multirow{4}{*}{LLaMA}
      & Baseline (NL)      & \meanstd{41.9}{3.8} \\
    &  & Constrained     & \textit{\meanstd{28}{12.2}} \\
    &  & NL-to-Format    & \textit{\meanstd{70.1}{5.3}} \\
    &  & In-Writing*     & \textbf{\meanstd{70.3}{3.8}} \\
    \cline{2-4}

    & \multirow{4}{*}{Gemma}
      & Baseline (NL)        &  \meanstd{9.3}{8.1} \\
    &  & Constrained     & \textit{\meanstd{39}{6.8}} \\
    &  & NL-to-Format    & \textit{\meanstd{56.8}{9.8}} \\
    &  & In-Writing*     & \textbf{\meanstd{58.4}{4.8}} \\
    \hline
    
    \multirow{8}{*}{Shuffled Obj}
    & \multirow{4}{*}{LLaMA}
      & Baseline (NL)        & \meanstd{1}{1.1} \\
    &  & Constrained     & \textit{\meanstd{15.7}{11}} \\
    &  & NL-to-Format    & \textit{\meanstd{27}{5.5}} \\
    &  & In-Writing*     & \textbf{\meanstd{39.2}{4.7}} \\
    \cline{2-4}

    & \multirow{4}{*}{Gemma}
      & Baseline (NL)        & \meanstd{10}{1.7} \\
    &  & Constrained     & \textit{\meanstd{50.5}{8.9}} \\
    &  & NL-to-Format    & \textit{\meanstd{49.4}{5.8}} \\
    &  & In-Writing*     & \textbf{\meanstd{50.5}{5.6}} \\
    \hline

  \end{tabular}

  \caption{Zero-shot results for GSM8K, Last Letter and ShuffledObj using LLaMA3-8B-it and Gemma2-9B-it, comparing Baseline (NL), Constrained decoding, NL-to-Format, and In-Writing*. Reported values are mean accuracy $\pm$ standard deviation over 9 variations. \textit{Italicized} results are reported by \citet{tam-etal-2024-speak}, while \textbf{bolded} results indicate the highest accuracy.}
  \label{tab:reasoning_base_lmsf_results}
\end{table}

We further evaluate on newer models of varying sizes, including Qwen3 (1.7B, 4B, 8B), Qwen3.5 (2B, 4B, 9B), and SmolLM3-3B, on three reasoning tasks (GSM8K, Last Letter Concatenation, and Shuffled Objects), which have been reported to exhibit notable performance degradation under constrained decoding, using the same few-shot examples as in prior work \citep{tam-etal-2024-speak}. The results are shown in Tables~\ref{tab:gsm8k_results}, \ref{tab:llc_results}, and \ref{tab:shuffleobj_results}.

\begin{figure}[t]
  \includegraphics[width=\columnwidth]{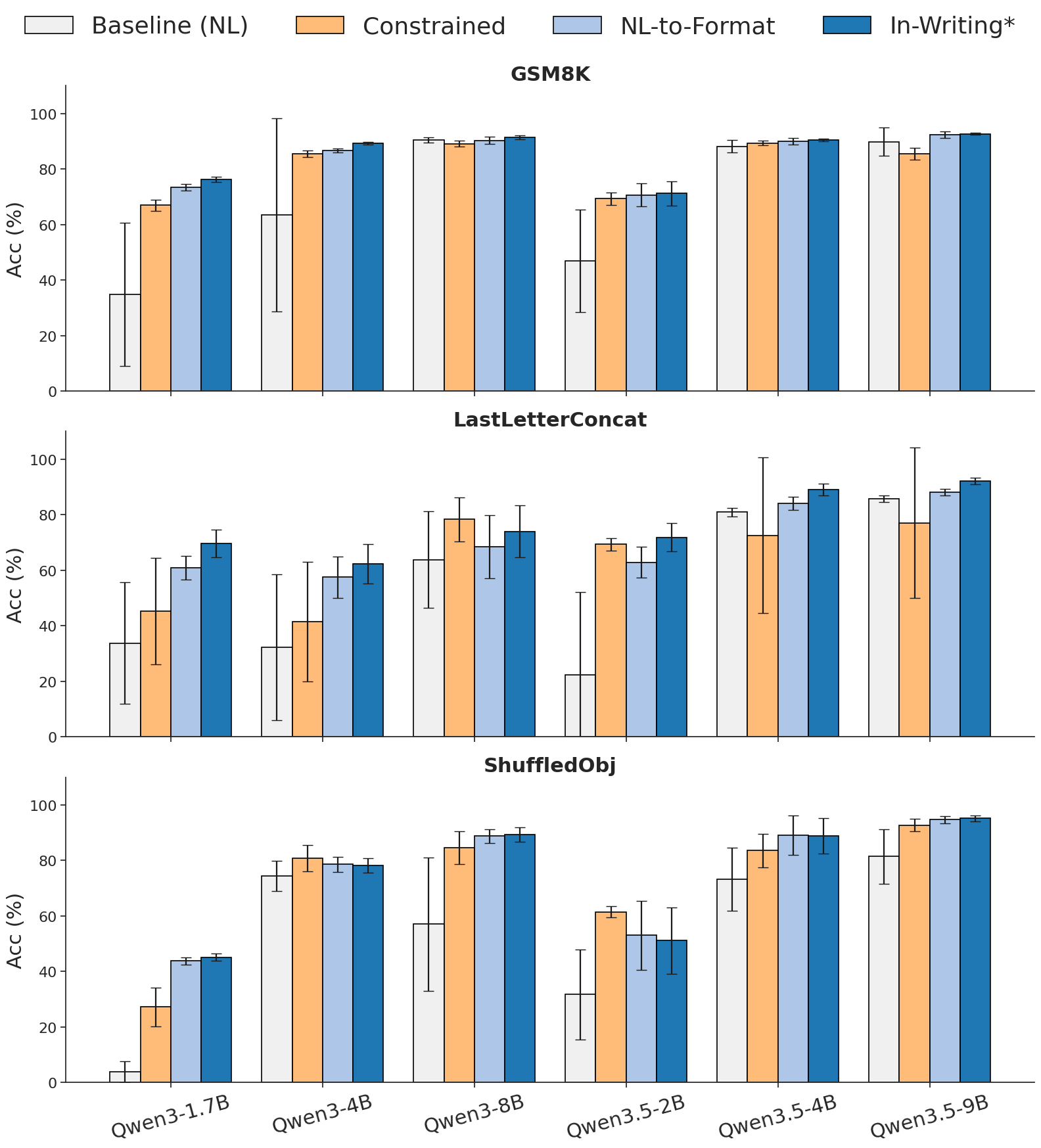}
    \caption{Zero-shot comparison for GSM8K, Last Letter, and ShuffledObj using various Qwen3 and Qwen3.5 model sizes, comparing Baseline (NL), Vanilla Constrained, NL-to-Format, and In-Writing*. In-Writing* performs best in nearly all settings.}
    \label{fig:qwen_lmsf}
\end{figure}

From Figure~\ref{fig:qwen_lmsf} and Tables~\ref{tab:reasoning_base_lmsf_results},~\ref{tab:gsm8k_results}, \ref{tab:llc_results}, \ref{tab:shuffleobj_results}, \ref{tab:classification_results}, \ref{tab:parse_results}, \ref{tab:overlap_results} and \ref{tab:token_results}, we make five observations:

\paragraph{(1) Consistency across extraction methods.}
Baseline (NL), NL-to-Format, and In-Writing* share identical reasoning traces; performance differences stem solely from answer extraction. We find that parser-based extraction consistently outperforms regex-based methods, while In-Writing* yields further gains of up to 27\% over LLM-based parsing.

\paragraph{(2) Parser dependence and limitations.}
NL-to-Format performance depends strongly on the parser (prompt or model) used in the second stage. While generally effective, parser quality varies across tasks and may fail (e.g., DDXPlus), leading to significant performance drops. See Appendix~\ref{sec:results-LMs} for details. It is also important to consider multi-field structured extraction, which is critical for industrial use cases and is not reliably followed by natural generation \citep{liu2024we}.

\paragraph{(3) Variance in constrained decoding.}
Vanilla constrained decoding exhibits high variance across tasks. Although larger and newer models reduce this gap, they still underperform other methods, consistent with prior work \citep{tam-etal-2024-speak, lee2026format}, leaving a substantial performance gap. In some cases, JSON-style constraints improve performance under identical prompts, but this gain is largely attributable to increased token generation.

\paragraph{(4) Premature triggering - a failure mode of constrained decoding.} In-Writing-Base results indicate that premature triggering truncates reasoning and reduces output quality, especially on mathematically intensive tasks such as GSM8K (over 30\% degradation compared to In-Writing*), whereas its impact on classification tasks is minimal.

\paragraph{(5) Effectiveness of In-Writing*.} By simply setting \texttt{<eos>} as the unique trigger token, In-Writing* mitigates premature triggering - a failure mode of constrained decoding. In-Writing* generally outperforms NL-to-Format and vanilla constrained decoding across model scales, achieves 100\% format validity without requiring larger models or explicit formatting guidance, and incurs only minimal overhead (5--20 tokens).

\subsubsection{Parsibility Analysis}

From Section~\ref{sec:full-experimental-results} and Table~\ref{tab:parse_results}, we observe that prompting the desired output format is necessary but not sufficient for robust parsing, as the NL baseline exhibits substantial variation in parse rates across model sizes.

Vanilla constrained decoding and NL-to-Format improve parse rates; however, performance still varies notably depending on token budget or parser model and prompt design.


In contrast, In-Writing achieves 100\% parse rate with low inference overhead, as only the formatting component requires constrained construction, while the reasoning segment remains unconstrained and is easily controlled via a maximum token limit. This makes the proposed framework computationally efficient while guaranteeing correct formatting.

\subsubsection{Overlap analysis between NL-to-Format and In-Writing}
\label{subsubsec:overlap_nl_format_inwriting}
With identical prompts, NL-to-Format and In-Writing* produce identical reasoning traces and differ only in the formatting step. This motivates analyzing the extraction capability of the two methods under the same reasoning process. We therefore introduce an overlap analysis with four metrics: joint success ($\cap$), where both methods correctly parse the answer; joint failure (Neither), where neither is correct; and two asymmetric cases, where only one method is correct.

\begin{figure}[t]
  \includegraphics[width=\columnwidth]{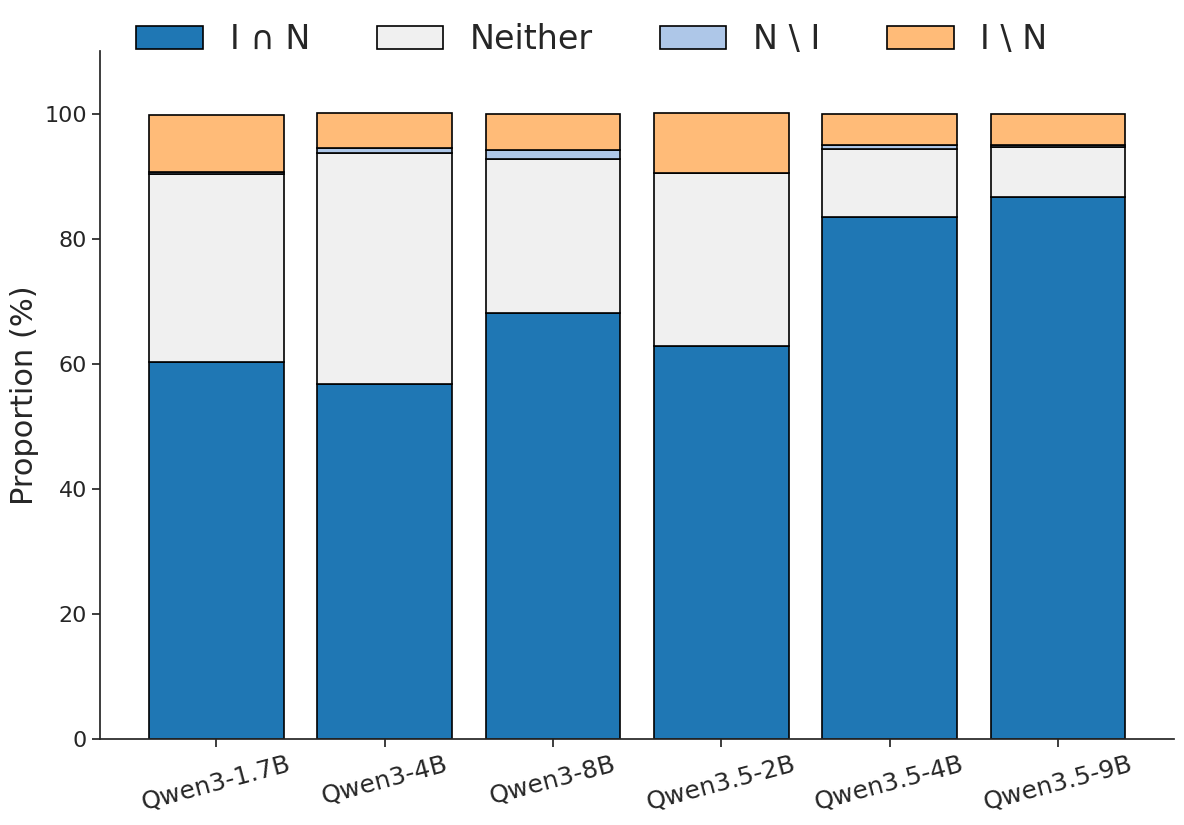}
    \caption{Overlap analysis on LastLetter between In-Writing* ($\mathbf{I}$) and NL-to-Format ($\mathbf{N}$), showing joint success ($\mathbf{I \cap N}$), joint failure (Neither), and asymmetric success cases ($\mathbf{N \setminus I}$, where NL-to-Format is correct and In-Writing* is incorrect; $\mathbf{I \setminus N}$, where In-Writing* is correct and NL-to-Format is incorrect). Values are reported in percentages. In–Writing* recovers most NL-to-Format failures, whereas the converse is far less common.}
    \label{fig:overlap_lastletter}
\end{figure}

From Figure~\ref{fig:overlap_lastletter} and Table~\ref{tab:overlap_results}, we observe that In-Writing* correctly parses many cases where NL-to-Format fails, whereas the reverse is much less common. It achieves up to an 11.8\% improvement, driven by both stronger parsing and an inherent correction capability (Figure~\ref{fig:natural_vs_structured}). This syntactic correction emerges from token masking induced by predefined constraints (Eq.~\ref{eq:InWriting_constrained_full_formula}). These results highlight In-Writing’s effectiveness for accurate, format-compliant extraction and correction without additional models or post-hoc processing. Further analysis is provided in Appendix~\ref{sec:results-LMs}.

\subsubsection{Token Efficiency Analysis}
As shown in Table~\ref{tab:token_results}, the number of newly generated tokens increases consistently from Baseline (NL) to NL-to-Format and In-Writing*. NL-to-Format adds 2–5 tokens, while In-Writing* adds 5–20 tokens. In contrast, vanilla constrained decoding is inconsistent, as it introduces JSON syntax tokens at the start of generation, thereby altering the reasoning trace.

There are two key differences between NL-to-Format and In-Writing*: (1) The parser in NL-to-Format only extracts the answer without enforcing structured format requirements (e.g., JSON or YAML), resulting in lower additional token cost; however, when format requirements are included, this cost becomes comparable to In-Writing*; (2) NL-to-Format relies on an additional LLM for parsing, which takes the parser prompt and the first model’s output (NL) as input, leading to substantially higher input token usage and computational overhead.

Overall, an additional 5–20 tokens (which scales with the complexity of the desired grammar) is a reasonable trade-off for improved format correctness and performance.

\subsection{CRANE Comparison: Limitations of Grammar-Constrained Decoding in LLM Reasoning}\label{sec:CRANE_comparison}

We evaluate In-Writing* on GSM-Symbolic using the same model families, and eight-shot setting as prior work \citep{banerjee2025crane}. Unlike the previous reasoning tasks, GSM-Symbolic requires models to generate symbolic solution expressions. Since the provided reference solutions are not always universally valid, we manually evaluate model outputs against the ground-truth formulae. Results are shown in Table~\ref{tab:gsmsym_results}. Evaluation details and additional discussion are provided in Appendix~\ref{sec:gsm-symbolic}.

CRANE constrains reasoning to grammar-restricted fields (\texttt{<< >>}), which can limit symbolic reasoning capability. In particular, restricting the set of valid operators prevents models from generating necessary functions such as \texttt{round()}, while also encouraging incorrect use of permitted operators such as \texttt{int()}. In contrast, In-Writing* allows unrestricted reasoning while constraining only final answer extraction. Under nearly identical prompts, it consistently outperforms CoT due to improved parsing success. Overall, our approach yields performance improvements of up to 32\% compared to CRANE.

\section{Conclusion}

We present In-Writing, a unified framework that decouples reasoning expressiveness from guaranteed output formatting within a single inference call, reframing constrained decoding not as a generation method but as a post-reasoning parser and corrector.
Our experiments across a wide range of reasoning and classification benchmarks show that the proposed approach improves performance, even with misaligned prompts, while maintaining comparable or lower computational overhead.
By simply designating \texttt{<eos>} as a unique trigger token, we show that \textit{premature triggering} can be mitigated, preventing disruption of the reasoning trace.

This work systematically studies In-Writing under a deliberately suboptimal prompting setup to evaluate its robustness. Future work should explore prompt optimization strategies that could further enhance its performance and stability, making it an even more effective framework for structured decoding.

\subsection*{Limitations}\label{sec:limitations-solutions}
A key limitation of this study is the lack of prompt optimization for In-Writing. We focus exclusively on a setting in which no explicit guidance is provided for the desired output format of In-Writing.

As shown in Table~\ref{tab:gsm8k_results} and Table~\ref{tab:llc_results}, increasing the number of few-shot examples improves Baseline performance but degrades both NL-to-Format and In-Writing. This effect is not attributable to differences in reasoning traces, which remain identical across methods, but rather to increased adherence to the \textbf{``answer is:''} prefix, which disproportionately benefits the baseline extraction mechanism.
Table~\ref{tab:overlap_results} further shows marginal cases where NL-to-Format succeeds while In-Writing fails. This is expected, as no prompting strategy is specifically tailored to In-Writing extraction. We anticipate that these limitations could be mitigated through more carefully designed prompts optimized for In-Writing.

\bibliography{custom}

\appendix
\section{Appendix}
\label{sec:appendix}

\subsection{Illustration of the In-Writing Framework}\label{sec:illustration_in_writing}

\begin{figure*}[h]
    \centering
    \includegraphics[width=0.8\textwidth]{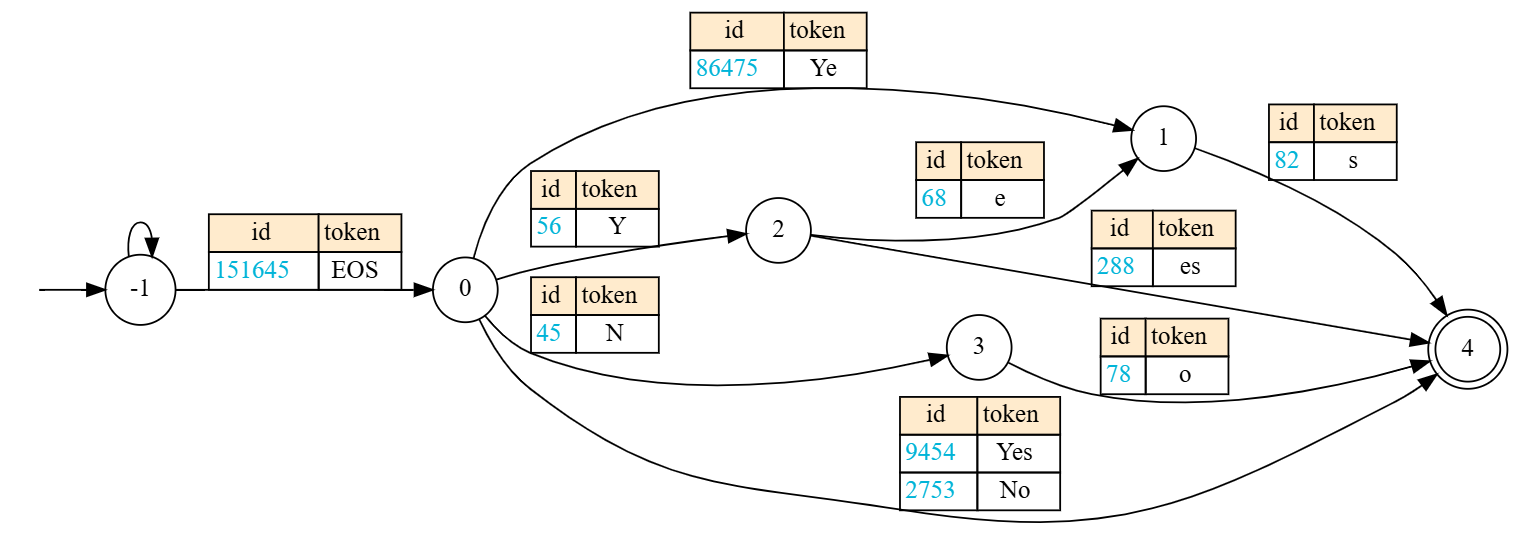}
    \caption{Illustrative example of the In-Writing approach. The model first generates unconstrained reasoning and then switches to guided decoding (state~$0$) once a trigger token is produced.}
    \label{fig:text2schema-example}
\end{figure*}

A simple In-Writing illustration is shown in Figure~\ref{fig:text2schema-example}, which demonstrates the process using a simple regex matching either ``Yes.'' or ``No.''. In standard constrained decoding \cite{willard2023efficient}, generation starts in state $0$ and is directly guided by an FSM that masks invalid tokens. In contrast, In-Writing performs free-form generation in state $-1$ and only switches to FSM-guided decoding after a trigger token, restricting outputs to regex- or grammar-constrained tokens. Tokenization may vary across models, but the underlying mechanism is unchanged.

\subsection{NL-to-Format Comparison: Experimental Setting Discussion}\label{sec:prompt-template}

\subsubsection{Prompt Templates and Experimental Settings for NL-to-Format Comparison Experiments}

For a direct comparison between the work of \citet{tam-etal-2024-speak} and our approach, we use the exact same prompt template and content. The base prompt template, in which all content is provided under the user role, is as follows:

\begin{quote}
\textbf{Follow the instruction to complete the task:} \texttt{\{task\_description\}}\\
\textbf{Instruct:} \texttt{\{format\_description\}}\\
\textbf{Few-shot examples:} \texttt{\{few\_shots\}}\\
\textbf{Question:} \texttt{\{question\}}
\end{quote}

\subsubsection{Prompt Inconsistencies in Previous Work}\label{sec:error-lmsf}

By investigating the outputs of the models provided in previous work \citep{tam-etal-2024-speak}, where constrained decoding was shown to harm LLM performance, we found several errors and inconsistencies. 
In the Shuffle Object Task Description Variations, task variation 1 and task variation 2 include seven answer choices, but the prompt incorrectly instructs the model to choose from only four choices: \\ \\
\noindent \textbf{Task Description Variation 1:}
\begin{tcolorbox}[colback=gray!10, colframe=black, boxrule=0.5pt, arc=2mm]
In this task, you are tasked to answer the following commonsense knowledge task. Read carefully each of the last questions and think step by step before answering. Make sure the answer only contains one of these \textbf{four} choices: A, B, C, D, E, F, G.
\end{tcolorbox}

\noindent \textbf{Task Description Variation 2:}
\begin{tcolorbox}[colback=gray!10, colframe=black, boxrule=0.5pt, arc=2mm]
Read carefully each of the last questions and think step by step before answering. Make sure the answer only contains one of these \textbf{four} choices: A, B, C, D, E, F, G. In this task, you are tasked to answer the following commonsense knowledge task.
\end{tcolorbox}

Format variations are used to instruct LLMs to reason first and then provide answers in a specific format. These variations are intended to be task-independent and should therefore remain consistent across tasks. However, in the previous work \citep{tam-etal-2024-speak}, the format variations are sometimes kept unchanged and sometimes vary across tasks, which raises concerns about evaluation consistency.

\begin{tcolorbox}[colback=gray!10, colframe=black, boxrule=0.5pt, arc=2mm]

\textbf{Format Variations for GSM8K and Last Letter Concatenation}

\textbf{Variation 1}
Instruct: Provide your output in the following text format:
Answer: <reasoning first>. The final answer is <answer>

\textbf{Variation 2}
Instruct: Provide your output in the following text format:
Step by step reasoning: ...
Answer: The final answer is ...

\textbf{Variation 3}
Instruct: Provide your output in the following text format:
Answer: <think step by step>. The final answer is <answer>

\end{tcolorbox}

\begin{tcolorbox}[colback=gray!10, colframe=black, boxrule=0.5pt, arc=2mm]

\textbf{Format Variations for Shuffled Objects}

\textbf{Variation 1}
Instruct: Now, take a deep breath and work on this problem step by step
to derive the most likely choice.
Provide your output in the following valid text format:
Answer: ...reasoning here... The answer is ...

\textbf{Variation 2}
Instruct: Provide your output in the following text format:
Step by step reasoning: ...
Answer: The final answer is ...

\textbf{Variation 3}
Instruct: Now, take a deep breath and work on this problem step by step
to derive the most likely answer.
Provide your output in the following valid text format:
Answer: [think step by step] The answer is [answer here]

\end{tcolorbox}

We also identify some missing or uninformative few-shot examples in the work of \citet{tam-etal-2024-speak}. For instance, in the Shuffled Object task, few-shot examples are not aligned with the target dataset. Examples include:
\begin{itemize}
    \item \textbf{Question}: ``Many people live in Ethiopia. The people are very thin and good at distance running.'' \\
    \textbf{Response}: \texttt{answer = ``race''}
    \item \textbf{Question}: ``The Norwegian man was boring.'' \\
    \textbf{Response}: \texttt{answer = ``race''}

\end{itemize}

We therefore restrict the evaluation of Shuffled Objects to the zero-shot prompting setting.

We use the same experimental settings as prior work \citep{tam-etal-2024-speak}: \texttt{seed=1}, \texttt{temperature=0}, and \texttt{do\_sample=False}, while default values are applied to all other unspecified parameters.

\subsection{CRANE Comparison: Experimental Setting Discussion}\label{sec:gsm-symbolic}
\subsubsection{Prompt Templates Used in CRANE Comparison Experiments}

We use the same Chain-of-Thought prompt as in prior work \citep{banerjee2025crane}, with a small instruction appended to the end of the prompt: \textit{Then Output the final answer as a symbolic expression in JSON format exactly like this: \texttt{"final\_answer": <symbolic expression>}}. Since the original prompt uses \texttt{<< >>} for reasoning, this encourages extraction of the final answer only. We keep the original few-shot examples unchanged.

\subsubsection{GSM-Symbolic dataset}\label{sec:gsm-symbolic-dataset-discussion}

By inspecting the GSM-Symbolic dataset used in prior work, we identify several inconsistencies between question requirements and gold answers, including operator misuse (e.g., \texttt{//} vs.\ \texttt{/} or \texttt{int()}) and missing required operations such as \texttt{round()}. In many cases, the ground-truth solutions rely on \texttt{//} or \texttt{int()} to force integer-valued expressions. However, using floating-point division with \texttt{/} is often more mathematically sound and aligns better with how LLMs naturally generate solutions. Moreover, when rounding is required, \texttt{round()}---which rounds to the nearest integer---is generally more appropriate than truncation via \texttt{int()} (see Examples~1 and~51).

We also identify ambiguities in several questions where substituting symbolic variables with concrete values does not preserve a well-defined notion of correctness (see Example~55).

\begin{tcolorbox}[colback=gray!10, colframe=black, boxrule=0.5pt, arc=2mm]

\textbf{Example 1}  

Question: A fog bank rolls in from the ocean to cover a city. It takes \{t\} minutes to cover every \{d\} miles of the city. If the city is \{y\} miles across from the oceanfront to the opposite inland edge, how many minutes will it take for the fog bank to cover the whole city?

Answer: \texttt{y//d*t}

\textbf{Example 51}  

Question: "\{name\} has a \{product\} stand at the \{location\}. He sells three kinds of \{product\}s: \{item1\}, \{item2\}, and \{item3\}. He usually sells \{item1\} for \{cur\}\{p1\} per \{unit\}, \{item2\} for \{cur\}\{p2\} per \{unit\}, and \{item3\} for \{cur\}\{p3\} per \{unit\}. \{name\} has no change today, so he has decided to round all his prices to the nearest dollar. If \{name\} sells \{n1\} \{unit\}s of \{item1\}, \{n2\} \{unit\}s of \{item2\}, and \{n3\} \{unit\}s of \{item3\}, how much will he make?

Answer: \texttt{n1*int(p1) + n2*int(p2) + n3*int(p3)}

\textbf{Example 55}  

Question: \{name\} is learning to write and decides to keep re-writing the \{alphabets\} until she knows it. She writes it in full \{n1\} time(s), writes \{frac\} of it once, then re-writes everything she has already written. How many letters has \{name\} written in total?

Answer: \texttt{2 * (\{alphabets\} * \{n1\} + \{alphabets\} * \{frac\})}

\end{tcolorbox}


\subsubsection{Evaluation of GSM-Symbolic Experimental Results}

Due to frequent operator misuse or omission in the ground-truth annotations (see~\ref{sec:gsm-symbolic-dataset-discussion}), we manually evaluate only the formulas parsed from LLM outputs against the gold answers. In this evaluation, we do not distinguish between semantically equivalent division operators (e.g., \texttt{/}, \texttt{//}, and \texttt{frac}), nor between rounding-related operations such as \texttt{int()} and \texttt{round()}.

In some cases, parsed formulas contain ambiguities between currency and numerical values (e.g., \{cur\}\{p1\}), where such expressions are interpreted as raw numerical inputs. We conservatively mark these cases as incorrect even when the overall structure is reasonable.

For example:

Gold Answer: \texttt{n1*int(p1) + n2*int(p2) + n3*int(p3)}

Parsed Answer: \texttt{n1*int(\{cur\}\{p1\}) + n2*int(\{cur\}\{p2\}) + n3*int(\{cur\}\{p3\})}


\subsection{Experimental Results and  Discussion}\label{sec:results-LMs}

\subsubsection{Full Experimental Results}\label{sec:full-experimental-results}
We first re-evaluate the experiments of \citet{tam-etal-2024-speak} using both the same prior models and additional extension models (Tables~\ref{tab:gsm8k_results}, \ref{tab:llc_results}, \ref{tab:shuffleobj_results}, and \ref{tab:classification_results}), along with parsability results (Table~\ref{tab:parse_results}). We then further analyze the overlap between the NL-to-Format approach from prior work and our method (Table~\ref{tab:overlap_results}), followed by a token-efficiency analysis (Table~\ref{tab:token_results}).

We also compare In-Writing with CRANE---another hybrid constrained decoding framework---in Table~\ref{tab:gsmsym_results}.

\subsubsection{Limitations of LLMs as Parsers}\label{sec:error-lmsf}
One limitation of using a larger LLM as a parser is its high computational cost and latency, which can be impractical \citep{tam-etal-2024-speak, lee2026format}. This raises the question of why a smaller model would be used at all if a larger model is available for parsing.

We also observe from the logs of \citet{tam-etal-2024-speak} that LLM-based parsers may \textbf{correct} model outputs rather than only parse them. In the LastLetter task, models often insert hyphens or whitespace between letters; removing these can incidentally turn an incorrect output into a correct one. In such cases, the parser effectively modifies the prediction rather than extracting it.

Finally, LLM-based parsing is prompt-dependent and can be unstable. In DDXPlus logs, parsers sometimes add conversational prefixes (e.g., ``the final answer that matches the given list of valid diagnoses is'') before outputting \textbf{acute dystonic reactions}, making the result harder to parse despite being correct.

\subsubsection{In-Writing Is Not a Perfect Framework}\label{sec:constrained-decoding-limitation-parsability}

As shown in Table~\ref{tab:overlap_results}, cases where In-Writing is correct but NL-to-Format is incorrect are substantially more frequent than the reverse, especially in the LastLetter task. This improvement comes not only from better parsing, but also from the auto-correcting effect of constrained decoding. By enforcing a predefined regex or grammar, constrained decoding prevents invalid tokens (e.g., whitespace between letters), encouraging models to avoid such errors during constrained generation.

Even so, hybrid constrained decoding is far from perfect. Although this method shows promising results compared to NL-to-Format, it still fails in a non-negligible number of cases. As shown in Table~\ref{tab:overlap_results}, In-Writing fails on certain examples where NL-to-Format correctly parses the output.

We also identify some examples from the GSM8K benchmark where constrained decoding fails to correctly map the final answer into the \texttt{final\_answer} field. Some of these examples are shown below.

\begin{lstlisting}[basicstyle=\ttfamily\tiny,breaklines=true,columns=fullflexible]
"sample_id": 63,
"gold_answer": "25000",
"model_output": "Answer: For the first 20 years, Marcy's annual pension remains the same at $50000. After 20 years, she becomes entitled to 5% of the value of the pension per year, which is 5% of $50000 = 0.05 x $50000 = $2,500. The final answer is $52,500.

{  "final_answer" :  "52"  }",
"parsed_answer": "52",

"sample_id": 125,
"gold_answer": "10",
"model_output": "Answer: Let's start by identifying the number of letters that Jennie has already stamped, which is 30. Since she puts stamps on one-third of the letters needing stamps, the number of letters needing stamps is three times the number of already-stamped letters. Therefore, the number of letters needing stamps is 3 x 30 = 90.

Since Jennie has a pile of 60 letters needing stamps, the number of letters already in the pile of already-stamped letters when Jennie began is 30 - (90 - 60) = 30 - 30 = 0. The final answer is 0.

{  "final_answer" :  "30"  }",
"parsed_answer": "30"
\end{lstlisting}

While we cannot provide a full statistical analysis due to human resource limitations, we would like to emphasize that constrained decoding is far from a perfect parser, and future work is needed to improve its semantic correctness and parsability.

\subsubsection{Empirical Contradictions to the CRANE Framework}\label{sec:CRANE_contradiction}

In the work of \citet{mirzadeh2025gsm}, the performance degradation of the same models from GSM8K to GSM-Symbolic is noticeable but relatively modest (up to approximately 10\%).

While CRANE proposes an interesting mechanism that allows LLMs to switch between natural and constrained generation via delimiter-based control tokens (\texttt{<< >>}), and is theoretically well-motivated \citep{banerjee2025crane}, its empirical results deviate from prior observations on GSM-Symbolic. For instance, with LLaMA3.1-8B, CRANE achieves 33\% accuracy, whereas an LLM-based parser evaluated on GSM8K with LLaMA3-8B reaches up to 70\%. Assuming LLaMA3.1-8B and LLaMA3-8B are broadly comparable in capability, this corresponds to an absolute degradation of approximately 37\% when moving from GSM8K to GSM-Symbolic, highlighting a substantial performance drop under grammar-constrained reasoning. 

In contrast, with In-Writing, we obtain 59\% accuracy, corresponding to an estimated drop of around 10\% under the same assumption, which is consistent with the degradation trend observed by \citet{mirzadeh2025gsm}.

This trend is consistent across model sizes, where In-Writing substantially recovers the lost performance. These findings raise questions about the robustness and practical applicability of the CRANE framework in symbolic reasoning settings.

\begin{table*}
  \centering
  \begin{tabular}{lclllll}
    \hline
    \textbf{Model} & \textbf{Shots} & \textbf{Baseline (NL)} & \textbf{Constrained} & \makecell{\textbf{NL-to-}\\\textbf{Format}} & \makecell{\textbf{In-Writing-}\\\textbf{Base}} & \textbf{In-Writing*} \\
    \hline

    \multirow{1}{*}{Llama3-8B}
    & 0 & \meanstd{66.2}{9.5} & \textit{\meanstd{48.9}{6.7}} & \textit{\meanstd{74.7}{0.6}} & \meanstd{77.9}{1.3} & \textbf{\meanstd{77.9}{1.3}} \\

    \hline

    \multirow{1}{*}{Gemma2-9B}
    & 0 & \meanstd{82.7}{7.2} & \textit{\meanstd{84.2}{3.7}} & \textit{\meanstd{86.5}{0.6}} & \meanstd{86.9}{0.4} & \textbf{\meanstd{86.9}{0.4}} \\

    \hline

    \multirow{3}{*}{Qwen3-1.7B}
    & 0 & \meanstd{34.8}{25.8} & \meanstd{66.9}{2} & \meanstd{73.7}{1.2} & \meanstd{59}{5.4} & \textbf{\meanstd{76.2}{0.9}} \\
    & 1 & \meanstd{70.4}{2.6} & \meanstd{69.1}{1.8} & \meanstd{71.1}{2.6} & \meanstd{65.8}{1.6} & \textbf{\meanstd{71.8}{2.7}} \\
    & 4 & \meanstd{72.9}{1.9} & \meanstd{69.5}{2.6} & \meanstd{73.2}{1.9} & \meanstd{69.3}{1.6} & \textbf{\meanstd{73.6}{2.3}} \\

    \hline

    \multirow{3}{*}{Qwen3-4B}
    & 0 & \meanstd{62.7}{34.7} & \meanstd{86.5}{1.2} & \meanstd{86.9}{0.7} & \meanstd{73.4}{8.1} & \textbf{\meanstd{88.8}{0.4}} \\
    & 1 & \meanstd{86.6}{0.8} & \meanstd{86.7}{1.1} & \meanstd{85.9}{1.1} & \meanstd{85.2}{1.2} & \textbf{\meanstd{86.7}{0.8}} \\
    & 4 & \meanstd{86.6}{0.8} & \textbf{\meanstd{87.5}{0.9}} & \meanstd{85.9}{1} & \meanstd{85.8}{0.9} & \meanstd{86.6}{0.9} \\

    \hline

    \multirow{3}{*}{Qwen3-8B}
    & 0 & \meanstd{91.7}{0.9} & \meanstd{89.2}{1} & \meanstd{89.8}{1.3} & \meanstd{79.1}{5.8} & \textbf{\meanstd{91.8}{0.7}} \\
    & 1 & \meanstd{89.1}{1.7} & \meanstd{90.1}{0.3} & \meanstd{89.2}{0.8} & \meanstd{89.4}{0.5} & \textbf{\meanstd{90.1}{0.5}} \\
    & 4 & \meanstd{89.5}{1.1} & \meanstd{85.5}{6.7} & \meanstd{89.1}{0.8} & \meanstd{89.2}{0.5} & \textbf{\meanstd{89.9}{0.3}} \\

    \hline

    \multirow{3}{*}{Qwen3.5-2B}
    & 0 & \meanstd{46.5}{18.4} & \meanstd{69.1}{2.2} & \meanstd{70.5}{4.1} & \meanstd{46.7}{6.4} & \textbf{\meanstd{71.4}{4.4}} \\
    & 1 & \meanstd{43.5}{11.4} & \meanstd{56.6}{1.7} & \textbf{\meanstd{57.8}{4.1}} & \meanstd{45.8}{3.2} & \meanstd{55.5}{5.6} \\
    & 4 & \meanstd{61.1}{6.3} & \meanstd{63.5}{1.1} & \meanstd{65.5}{2.1} & \meanstd{57}{2.2} & \textbf{\meanstd{65.8}{2}} \\

    \hline

    \multirow{3}{*}{Qwen3.5-4B}
    & 0 & \meanstd{88.1}{2.3} & \meanstd{88.6}{0.9} & \meanstd{90.2}{1.1} & \meanstd{65.8}{9.5} & \textbf{\meanstd{90.7}{0.5}} \\
    & 1 & \meanstd{87}{1.2} & \meanstd{86.5}{0.6} & \meanstd{88.1}{1.5} & \meanstd{73.9}{6.1} & \textbf{\meanstd{88.3}{1.1}} \\
    & 4 & \meanstd{88.5}{1.4} & \meanstd{88}{0.7} & \meanstd{89.6}{1.1} & \meanstd{72.6}{8.7} & \textbf{\meanstd{90}{0.6}} \\

    \hline

    \multirow{3}{*}{Qwen3.5-9B}
    & 0 & \meanstd{89.9}{5.1} & \meanstd{85.7}{2.2} & \meanstd{92.1}{1.2} & \meanstd{61.9}{3.4} & \textbf{\meanstd{93.4}{0.3}} \\
    & 1 & \meanstd{90.3}{0.3} & \meanstd{87.5}{1.8} & \meanstd{90.1}{1} & \meanstd{79.7}{1.6} & \textbf{\meanstd{90.9}{0.4}} \\
    & 4 & \meanstd{91.8}{0.4} & \meanstd{88.9}{1.1} & \meanstd{91.7}{0.5} & \meanstd{80.9}{2.4} & \textbf{\meanstd{92.4}{0.3}} \\

    \hline

    \multirow{3}{*}{SmolLM3-3B}
    & 0 & \meanstd{72.8}{11.8} & \meanstd{46.7}{11.4} & \meanstd{81.8}{1.2} & \meanstd{69}{15.4} &  \textbf{\meanstd{83}{1.1}} \\
    & 1 & \meanstd{76.1}{8} & \meanstd{63.1}{13.2} & \meanstd{80.5}{0.9} & \meanstd{70.9}{11.8} & \textbf{\meanstd{81.6}{0.9}} \\
    & 4 & \meanstd{76.2}{8.7} & \meanstd{63.1}{12.2} & \meanstd{81.1}{0.8} & \meanstd{71}{13.8} & \textbf{\meanstd{82.1}{0.6}} \\

    \hline

  \end{tabular}
  \caption{\label{tab:gsm8k_results}
    Results on GSM8K grouped by model across zero-, one-, and four-shot settings. Reported values are mean accuracy $\pm$ standard deviation over 9 variations. \textit{Italicized} results are reported by \citet{tam-etal-2024-speak}, while \textbf{bolded} results indicate the highest accuracy
  }
\end{table*}

\begin{table*}
  \centering
  \begin{tabular}{lclllll}
    \hline
    \textbf{Model} & \textbf{Shots} & \textbf{Baseline (NL)} & \textbf{Constrained} & \makecell{\textbf{NL-to-}\\\textbf{Format}} & \makecell{\textbf{In-Writing-}\\\textbf{Base}} & \textbf{In-Writing*} \\
    \hline

    \multirow{1}{*}{Llama3-8B}
    & 0 & \meanstd{41.9}{3.8} & \textit{\meanstd{28}{12.2}} & \textit{\meanstd{70.1}{5.3}} & \meanstd{70}{4} & \textbf{\meanstd{70.3}{3.8}} \\

    \hline

    \multirow{1}{*}{Gemma2-9B}
    & 0 & \meanstd{9.3}{8.1} & \textit{\meanstd{39}{6.8}} & \textit{\meanstd{56.8}{9.8}} & \meanstd{58.3}{4.8} & \textbf{\meanstd{58.4}{4.8}} \\
    \hline

    \multirow{3}{*}{Qwen3-1.7B}
    & 0 & \meanstd{33.6}{21.9} & \meanstd{45.4}{19.1} & \meanstd{60.7}{4.3} & \meanstd{69.2}{4.6} & \textbf{\meanstd{69.7}{4.9}} \\
    & 1 & \meanstd{54.1}{4.6} & \meanstd{35.8}{24.6} & \meanstd{58}{3.1} & \meanstd{66.2}{3.1} & \textbf{\meanstd{66.2}{3.1}} \\
    & 4 & \meanstd{55}{3.4} & \meanstd{39}{19.7} & \meanstd{56.7}{3.5} & \meanstd{65.8}{3.7} & \textbf{\meanstd{65.9}{3.3}} \\

    \hline

    \multirow{3}{*}{Qwen3-4B}
    & 0 & \meanstd{32.1}{26.2} & \meanstd{41.9}{21.6} & \meanstd{57.5}{7.5} & \meanstd{61.9}{6.9} & \textbf{\meanstd{62.2}{7.1}} \\
    & 1 & \meanstd{60.1}{10.5} & \meanstd{21.1}{26.6} & \meanstd{67.6}{4.7} & \meanstd{73}{3.9} & \textbf{\meanstd{73.3}{3.8}} \\
    & 4 & \meanstd{75.5}{2.9} & \meanstd{27.6}{31} & \meanstd{76.4}{3.4} & \meanstd{80.8}{3} & \textbf{\meanstd{80.9}{2.9}} \\

    \hline

    \multirow{3}{*}{Qwen3-8B}
    & 0 & \meanstd{63.8}{17.4} & \textbf{\meanstd{78.4}{7.9}} & \meanstd{69.4}{11.4} & \meanstd{74}{9.4} & \meanstd{74}{9.4} \\
    & 1 & \meanstd{65.1}{8.4} & \meanstd{64.4}{20.3} & \meanstd{67.2}{6.7} & \meanstd{70.6}{5.9} & \textbf{\meanstd{70.6}{5.9}} \\
    & 4 & \meanstd{72.7}{2.4} & \textbf{\meanstd{77.5}{3.4}} & \meanstd{73.9}{2.3} & \meanstd{76}{2.1} & \meanstd{76}{2.1} \\

    \hline

    \multirow{3}{*}{Qwen3.5-2B}
    & 0 & \meanstd{22.4}{29.9} & \meanstd{69.1}{2.2} & \meanstd{62.9}{5.6} & \meanstd{71.9}{4.9} & \textbf{\meanstd{72.4}{5.1}} \\
    & 1 & \meanstd{17.4}{22.4} & \textbf{\meanstd{56.6}{1.7}} & \meanstd{39.6}{19} & \meanstd{47.3}{21.7} & \meanstd{47.4}{21.8} \\
    & 4 & \meanstd{58.7}{2.7} & \meanstd{63.5}{1.1} & \meanstd{60.1}{3.1} & \meanstd{69.1}{2.2} & \textbf{\meanstd{69.3}{2.4}} \\

    \hline

    \multirow{3}{*}{Qwen3.5-4B}
    & 0 & \meanstd{81.6}{1.6} & \meanstd{71.6}{28} & \meanstd{84.1}{2.3} & \meanstd{87.3}{2.1} & \textbf{\meanstd{88.4}{2.1}} \\
    & 1 & \meanstd{81.9}{2.3} & \meanstd{63.2}{35.1} & \meanstd{82.7}{1.9} & \meanstd{87.4}{1.8} & \textbf{\meanstd{88.2}{2.4}} \\
    & 4 & \meanstd{82}{1.4} & \meanstd{63.8}{34.4} & \meanstd{83.4}{1.1} & \meanstd{87.5}{2.1} & \textbf{\meanstd{88.1}{1.7}} \\

    \hline

    \multirow{3}{*}{Qwen3.5-9B}
    & 0 & \meanstd{85.9}{1.2} & \meanstd{77}{27} & \meanstd{86.8}{1.2} & \meanstd{89.9}{1.6} & \textbf{\meanstd{91.7}{1.2}} \\
    & 1 & \meanstd{85.5}{1.9} & \meanstd{81.3}{27.3} & \meanstd{86}{1.5} & \meanstd{89.4}{1.3}& \textbf{\meanstd{90.3}{1.8}} \\
    & 4 & \meanstd{87.8}{0.9} & \meanstd{82.1}{26.2} & \meanstd{88.7}{1.6} & \meanstd{90.8}{1} & \textbf{\meanstd{92}{0.8}} \\

    \hline

    \multirow{3}{*}{SmolLM3-3B}
    & 0 & \meanstd{21.7}{21.2} & \meanstd{13.9}{18.6} & \meanstd{54.2}{8.3} & \meanstd{64.1}{9.4} & \textbf{\meanstd{65.3}{9.7}} \\
    & 1 & \meanstd{43.9}{13.1} & \meanstd{8.7}{14.7} & \meanstd{55.5}{7.4} & \meanstd{64}{6.1} & \textbf{\meanstd{65.6}{6.5}} \\
    & 4 & \meanstd{45.1}{9.3} & \meanstd{7.9}{18.5} & \meanstd{54.2}{5.6} & \meanstd{60.6}{5.5} & \textbf{\meanstd{63}{5.5}} \\

    \hline

  \end{tabular}
  \caption{\label{tab:llc_results}
    Results on Last Letter Concatenation grouped by model across zero-, one-, and four-shot settings. Reported values are mean accuracy $\pm$ standard deviation over 9 variations. \textit{Italicized} results are reported by \citet{tam-etal-2024-speak}, while \textbf{bolded} results indicate the highest accuracy
  }
\end{table*}

\begin{table*}
  \centering
  \begin{tabular}{llllll}
    \hline
    \textbf{Model} & \textbf{Baseline (NL)} & \textbf{Constrained} & \makecell{\textbf{NL-to-}\\\textbf{Format}} & \makecell{\textbf{In-Writing-}\\\textbf{Base}} & \textbf{In-Writing*} \\
    \hline

    Llama3-8B              & \meanstd{1}{1.1} & \textit{\meanstd{15.7}{11}} & \textit{\meanstd{27.0}{5.5}}        & \meanstd{39.2}{4.7} & \textbf{\meanstd{39.2}{4.7}} \\
    Gemma2-9B              & \meanstd{10}{7} & \textit{\meanstd{50.5}{8.9}} & \textit{\meanstd{49.4}{5.8}}        & \meanstd{50.5}{5.6} & \textbf{\meanstd{50.5}{5.6}} \\

    Qwen3-1.7B         & \meanstd{3.5}{3.8} & \meanstd{27.7}{7} & \meanstd{44.4}{1.3}                        & \meanstd{43.4}{1.4} & \textbf{\meanstd{44.7}{1.3}} \\
    Qwen3-4B           & \meanstd{74.8}{5.4} & \textbf{\meanstd{81.1}{4.8}} &  \meanstd{78.4}{2.7}                         & \meanstd{78.3}{2.8} & \meanstd{78.8}{2.6} \\
    Qwen3-8B           & \meanstd{56.4}{24} & \meanstd{84.6}{5.9} & \meanstd{89.2}{2.5}                         & \meanstd{87.6}{2.7} & \textbf{\meanstd{89.2}{2.5}} \\
    Qwen3.5-2B         & \meanstd{32.5}{16.2} & \textbf{\meanstd{61.7}{2}} & \meanstd{52.3}{12.4}                         & \meanstd{46.3}{12.3} & \meanstd{50.5}{11.9} \\
    Qwen3.5-4B         & \meanstd{73}{11.4} & \meanstd{84.3}{6.1} & \meanstd{89.5}{7.1}                          & \meanstd{85}{9.3}  & \textbf{\meanstd{89.7}{6.4}} \\
    Qwen3.5-9B         & \meanstd{80.3}{9.8} & \meanstd{92.7}{2.2} & \meanstd{94.9}{1.3}                         & \meanstd{93.5}{2.1}  & \textbf{\meanstd{95.1}{1.1}} \\
    SmolLM3-3B         & \meanstd{3.7}{4.6} & \meanstd{30.4}{4.2} & \meanstd{34.9}{11.6}                         &  \meanstd{32.1}{12.7} & \textbf{\meanstd{35.3}{11.2}} \\

    \hline
  \end{tabular}
  \caption{\label{tab:shuffleobj_results}
    Results on Shuffled Objects grouped by model in the zero-shot setting. Reported values are mean accuracy $\pm$ standard deviation over 9 variations. \textit{Italicized} results are reported by \citet{tam-etal-2024-speak}, while \textbf{bolded} results indicate the highest accuracy.
    }
  
\end{table*}

\begin{table*}
  \centering
  \begin{tabular}{llllll}
    \hline
    \textbf{Task} & \textbf{Model} & \textbf{Constrained} & \makecell{\textbf{NL-to-}\\\textbf{Format}} & \makecell{\textbf{In-Writing-}\\\textbf{Base}} & \textbf{In-Writing*} \\
    \hline

    \multirow{2}{*}{MultiFin}
    & Llama3-8B & \textit{\meanstd{57.7}{2}} & \textit{\meanstd{60.3}{1.4}} & \meanstd{64.5}{3} & \textbf{\meanstd{64.5}{3}} \\
    & Gemma2-9B & \textbf{\textit{\meanstd{70.2}{0.7}}} & \textit{\meanstd{70}{0.4}} & \meanstd{70}{0.6} & \meanstd{70}{0.6} \\

    \hline

    \multirow{2}{*}{Sports}
    & Llama3-8B & \textit{\meanstd{73.4}{3.5}} & \textit{\meanstd{69.5}{12.7}} & \meanstd{77.4}{4.3} &  \textbf{\meanstd{77.4}{4.3}} \\
    & Gemma2-9B & \textit{\meanstd{72.7}{1.6}} & \textit{\meanstd{76.1}{2.3}} & \meanstd{76.2}{1.8} & \textbf{\meanstd{76.2}{1.8}} \\

    \hline

    \multirow{2}{*}{Task280}
    & Llama3-8B & \textit{\meanstd{39.5}{22.4}} & \textit{\meanstd{65.3}{3.4}} & \meanstd{74}{5.4} & \textbf{\meanstd{74}{5.4}} \\
    & Gemma2-9B & \textit{\meanstd{65.6}{11.7}} & \textit{\meanstd{69.8}{7.7}} & \meanstd{74.5}{7.4} & \textbf{\meanstd{74.5}{7.4}}\\

    \hline

    \multirow{2}{*}{DDXPlus}
    & Llama3-8B & \textit{\meanstd{23.4}{0.7}} & \textit{\meanstd{12}{15.2}} & \meanstd{30}{2.9} & \textbf{\meanstd{30.6}{3.1}} \\
    & Gemma2-9B & \textit{\meanstd{53}{0.2}} & \textit{\meanstd{22.9}{5.8}} & \meanstd{50.1}{1.6} &  \textbf{\meanstd{50.1}{1.6}} \\

    \hline
  \end{tabular}
  \caption{\label{tab:classification_results}
    Zero-shot prompting results for LLaMA3-8B-it and Gemma2-9B-it, averaged over 9 variations of each classification task. Reported values denote mean accuracy $\pm$ standard deviation across variations. \textit{Italicized} results are reported by \citet{tam-etal-2024-speak}, while \textbf{bolded} values indicate the highest accuracy.
  }
\end{table*}

\begin{table*}[t]
\centering
\begin{tabular}{llcccc}
\hline
\textbf{Task} & \textbf{Model} & \textbf{Baseline (NL)} & \textbf{Constrained} & \textbf{NL-to-Format} & \textbf{In-Writing*} \\
\hline

\multirow{6}{*}{GSM8K}
& Qwen3-1.7B 
& 47 
& 98 
& 100 
& \textbf{100} \\

& Qwen3-4B 
& 71 
& 99 
& 100
& \textbf{100} \\

& Qwen3-8B 
& 100 
& 99 
& 100
& \textbf{100} \\

& Qwen3.5-2B 
& 62
& 85 
& 100 
& \textbf{100} \\

& Qwen3.5-4B 
& 95
& 96 
& 100
& \textbf{100} \\

& Qwen3.5-9B 
& 98
& 92 
& 100
& \textbf{100} \\

\hline

\multirow{6}{*}{LastLetter}
& Qwen3-1.7B 
& 77 
& 100 
& 87 
& \textbf{100} \\

& Qwen3-4B 
& 53
& 99 
& 96 
& \textbf{100} \\

& Qwen3-8B 
& 96 
& 100 
& 95 
& \textbf{100} \\

& Qwen3.5-2B 
& 36
& 98 
& 90 
& \textbf{100} \\

& Qwen3.5-4B 
& 99 
& 100 
& 96
& \textbf{100} \\

& Qwen3.5-9B 
& 98 
& 100
& 96
& \textbf{100} \\

\hline

\multirow{6}{*}{Shuffled Obj}
& Qwen3-1.7B 
& 22 
& 99 
& 100 
& \textbf{100} \\

& Qwen3-4B 
& 97
& 99 
& 100 
& \textbf{100} \\

& Qwen3-8B 
& 68 
& 99 
& 100 
& \textbf{100} \\

& Qwen3.5-2B 
& 64 
& 80 
& 99 
& \textbf{100} \\

& Qwen3.5-4B 
& 95 
& 92 
& 100 
& \textbf{100} \\

& Qwen3.5-9B 
& 98 
& 93
& 100 
& \textbf{100} \\

\hline
\end{tabular}

\caption{\label{tab:parse_results}
Zero-shot prompting parsability results (in percentage), averaged over 9 prompt variations per reasoning task. Bolded values indicate the highest parse rate. In-Writing* achieved a perfect 100\% parse rate.
}
\end{table*}

\begin{table*}[t]
\centering
\begin{tabular}{ll|ll|ll}
\hline
\textbf{Task} & \textbf{Model} 
& $\mathbf{I \cap N}$ 
& \textbf{Neither} 
& $\mathbf{N \setminus I}$ 
& $\mathbf{I \setminus N}$ \\
\hline

\multirow{7}{*}{GSM8K}
& Qwen3-1.7B 
& 73.4 & 23.5 & 0.2 & \textbf{2.8} \\

& Qwen3-4B 
& 86.5 & 10.7 & 0.4 & \textbf{2.4} \\

& Qwen3-8B 
& 89.6 & 8.1 & 0.2 & \textbf{2.1} \\

& Qwen3.5-2B 
& 68.6 & 26.7 & 1.9 & \textbf{2.8} \\

& Qwen3.5-4B 
& 89.5 & 8.6 & 0.7 & \textbf{1.2} \\

& Qwen3.5-9B 
& 91.6 & 6.1 & 0.5 & \textbf{1.8} \\

& SmolLM3-3B 
& 81 & 16.2 & 0.8 & \textbf{2} \\

\hline

\multirow{7}{*}{LastLetter}
& Qwen3-1.7B 
& 60.3 & 30 & 0.4 & \textbf{9.1} \\

& Qwen3-4B 
& 56.7 & 37 & 0.8 & \textbf{5.6} \\

& Qwen3-8B 
& 68.1 & 24.7 & 1.3 & \textbf{5.9} \\

& Qwen3.5-2B 
& 62.9 & 27.6 & 0 & \textbf{9.6} \\

& Qwen3.5-4B 
& 83.4 & 10.9 & 0.7 & \textbf{5} \\

& Qwen3.5-9B 
& 86.6 & 8.1 & 0.2 & \textbf{5.1} \\

& SmolLM3-3B 
& 53.5 & 34 & 0.7 & \textbf{11.8} \\

\hline

\multirow{7}{*}{Shuffled Obj}
& Qwen3-1.7B 
& 43.2 & 54.1 & 1.2 & \textbf{1.5} \\

& Qwen3-4B 
& 78 & 20.8 & 0.4 & \textbf{0.8} \\

& Qwen3-8B 
& 88.9 & 10.3 & 0.4 & \textbf{0.4} \\

& Qwen3.5-2B 
& 48.4 & 45.5 & \textbf{3.9} & 2.1 \\

& Qwen3.5-4B 
& 86.3 & 7 & 3.2 & \textbf{3.5} \\

& Qwen3.5-9B 
& 94.2 & 4.2 & 0.7 & \textbf{0.9} \\

& SmolLM3-3B 
& 33.8 & 63.6 & 1.1 & \textbf{1.5} \\

\hline
\end{tabular}

\caption{\label{tab:overlap_results}
Overlap analysis between In-Writing* ($\mathbf{I}$) and NL-to-Format ($\mathbf{N}$). $\mathbf{I \cap N}$ denotes cases where both In-Writing* and NL-to-Format are correct, \textbf{Neither} indicates both are incorrect, $\mathbf{N \setminus I}$ corresponds to cases where NL-to-Format is correct but In-Writing* is incorrect, and $\mathbf{I \setminus N}$ corresponds to cases where In-Writing* is correct but NL-to-Format is incorrect. Values are reported as percentages. \textbf{Bolded} values indicate the method with larger asymmetric gain, i.e., the method that correctly solves more instances that the other method fails to solve.
}
\end{table*}

\begin{table*}[t]
\centering
\begin{tabular}{llcccc}
\hline
\textbf{Task} & \textbf{Model} & \textbf{Baseline (NL)} & \textbf{Constrained} & \textbf{NL-to-Format} & \textbf{In-Writing*} \\
\hline

\multirow{6}{*}{GSM8K}
& Qwen3-1.7B 
& 182.8 
& 168.6 
& 186.3 
& 195.2 \\

& Qwen3-4B 
& 153.9 
& 172.5 
& 157.3
& 172.6 \\

& Qwen3-8B 
& 167 
& 182 
& 170.4
& 187.3 \\

& Qwen3.5-2B 
& 294.7
& 316.6 
& 300 
& 305.7 \\

& Qwen3.5-4B 
& 262.4 
& 225.7 
& 265.7
& 274.2 \\

& Qwen3.5-9B 
& 252
& 270.1 
& 255.4
& 262.3 \\

\hline

\multirow{6}{*}{LastLetter}
& Qwen3-1.7B 
& 144.6 
& 141.8 
& 147.8 
& 153.9 \\

& Qwen3-4B 
& 100.9
& 104.1 
& 104 
& 114.8 \\

& Qwen3-8B 
& 122.1 
& 140.6 
& 125.2 
& 136.2 \\

& Qwen3.5-2B 
& 147.4
& 194.7 
& 151 
& 159.1 \\

& Qwen3.5-4B 
& 184.9 
& 174.2 
& 188
& 199.1 \\

& Qwen3.5-9B 
& 159.9 
& 167.9 
& 163
& 168.6 \\

\hline

\multirow{6}{*}{Shuffled Obj}
& Qwen3-1.7B 
& 286.6 
& 140.9 
& 288.6 
& 298.7 \\

& Qwen3-4B 
& 216.2 
& 221.1 
& 218.2 
& 228.8 \\

& Qwen3-8B 
& 279 
& 231.5 
& 281 
& 293.6 \\

& Qwen3.5-2B 
& 320.8 
& 359.5 
& 322.9 
& 310.5 \\

& Qwen3.5-4B 
& 400.6 
& 368.3 
& 402.7 
& 401.3 \\

& Qwen3.5-9B 
& 372.8 
& 367
& 374.9 
& 376.8 \\

\hline
\end{tabular}

\caption{\label{tab:token_results}
Comparison of generated output token counts, averaged over 9 prompt variations per reasoning task. In-Writing* introduces an additional 5--20 formatting-related tokens compared to the Baseline (NL).
}
\end{table*}

\begin{table*}
  \centering
  \begin{tabular}{lllllll}
    \hline
    \textbf{Model} & \textbf{Baseline} & \textbf{Constrained} & \textbf{CoT} & \textbf{CRANE} & \textbf{In-Writing*} \\
    \hline
    Qwen2.5-1.5B              & \textit{21} & \textit{22} & \textit{26} & \textit{31} &  \textbf{37} \\
    Qwen2.5-Coder-7B          & \textit{36} & \textit{35} & \textit{37} & \textit{39} & \textbf{69} \\

    Qwen2.5-Math-7B           & \textit{27} & \textit{29} & \textit{29} & \textit{38} & \textbf{47} \\
    Qwen2.5-Coder-14B         & \textit{42} & \textit{42} & \textit{42} & \textit{45} & \textbf{77} \\
    Llama3.1-8B               & \textit{21} & \textit{26} & \textit{30} & \textit{33} & \textbf{59} \\
    DeepSeek-R1-Qwen-7B       & \textit{18} & \textit{20} & \textit{24} & \textit{29} & \textbf{47} \\
    DeepSeek-R1-Llama-8B      & \textit{12} & \textit{13} & \textit{21} & \textbf{\textit{31}} & 30 \\
    
    DeepSeek-R1-Qwen-14B      & \textit{29} & \textit{30} & 32 & \textit{38} & \textbf{65} \\

    \hline
  \end{tabular}
  \caption{\label{tab:gsmsym_results}
    Eight-shot results on GSM-Symbolic task. \textit{Italicized} results are reported by \citet{banerjee2025crane}, while \textbf{bolded} values indicate the highest accuracy.
  }
\end{table*}

\end{document}